\documentclass[runningheads]{llncs}

\usepackage{eccv}

\usepackage{eccvabbrv}

\usepackage{graphicx}
\usepackage{booktabs}

\usepackage[accsupp]{axessibility}  

\usepackage{lipsum}
\usepackage{amssymb}
\usepackage{multirow}
\usepackage[<options>]{animate}
\usepackage{graphicx}
\usepackage[symbol]{footmisc}

\usepackage[pagebackref,breaklinks,colorlinks,citecolor=eccvblue]{hyperref}

\usepackage{orcidlink}

\usepackage{pifont}
\usepackage{bbding}

\usepackage{amsmath,amsfonts,bm}

\def\eqref#1{equation~\ref{#1}}

\def\1{\bm{1}}

\def\vx{{\bm{x}}}

\def\vz{{\bm{z}}}

\def\mK{{\bm{K}}}

\def\mQ{{\bm{Q}}}

\def\mV{{\bm{V}}}
\def\mW{{\bm{W}}}

\DeclareMathAlphabet{\mathsfit}{\encodingdefault}{\sfdefault}{m}{sl}
\SetMathAlphabet{\mathsfit}{bold}{\encodingdefault}{\sfdefault}{bx}{n}

\newcommand{\encoder}{\mathcal{E}}
\newcommand{\decoder}{\mathcal{D}}

\newcommand{\ignorethis}[1]{}

\makeatletter
\DeclareRobustCommand\onedot{\futurelet\@let@token\@onedot}
\def\@onedot{\ifx\@let@token.\else.\null\fi\xspace}

\def\eg{\emph{e.g}\onedot} 
\def\ie{\emph{i.e}\onedot}

\makeatother

\newrobustcmd{\B}{\bfseries}
\newcommand*{\rom}[1]{\expandafter\romannumeral #1}

\definecolor{mydarkblue}{rgb}{0,0.08,1}
\definecolor{mydarkgreen}{rgb}{0.02,0.6,0.02}
\definecolor{mydarkred}{rgb}{0.8,0.02,0.02}
\definecolor{mydarkorange}{rgb}{0.40,0.2,0.02}
\definecolor{mypurple}{RGB}{111,0,255}
\definecolor{myred}{rgb}{1.0,0.0,0.0}
\definecolor{mygold}{rgb}{0.75,0.6,0.12}
\definecolor{myblue}{rgb}{0,0.2,0.8}
\definecolor{mydarkgray}{rgb}{0.66,0.66,0.66}

\begin{document}

\title{VideoElevator: Elevating Video Generation Quality with Versatile Text-to-Image Diffusion Models
}

\titlerunning{VideoElevator}

\author{Yabo Zhang\inst{1} \and
Yuxiang Wei\inst{1} \and
Xianhui Lin \and
Zheng Hui \and
Peiran Ren \and \\
Xuansong Xie \and
Xiangyang Ji\inst{2} \and
Wangmeng Zuo\inst{1}$^{(}$\Envelope$^)$}

\authorrunning{Y. Zhang et al.}

\institute{Harbin Institute of Technology \and
Tsinghua University\\
\url{https://videoelevator.github.io}
}

\maketitle

\begin{center}
\animategraphics[autoplay,loop,width=0.99
\textwidth, trim=5 0 5 0, clip]{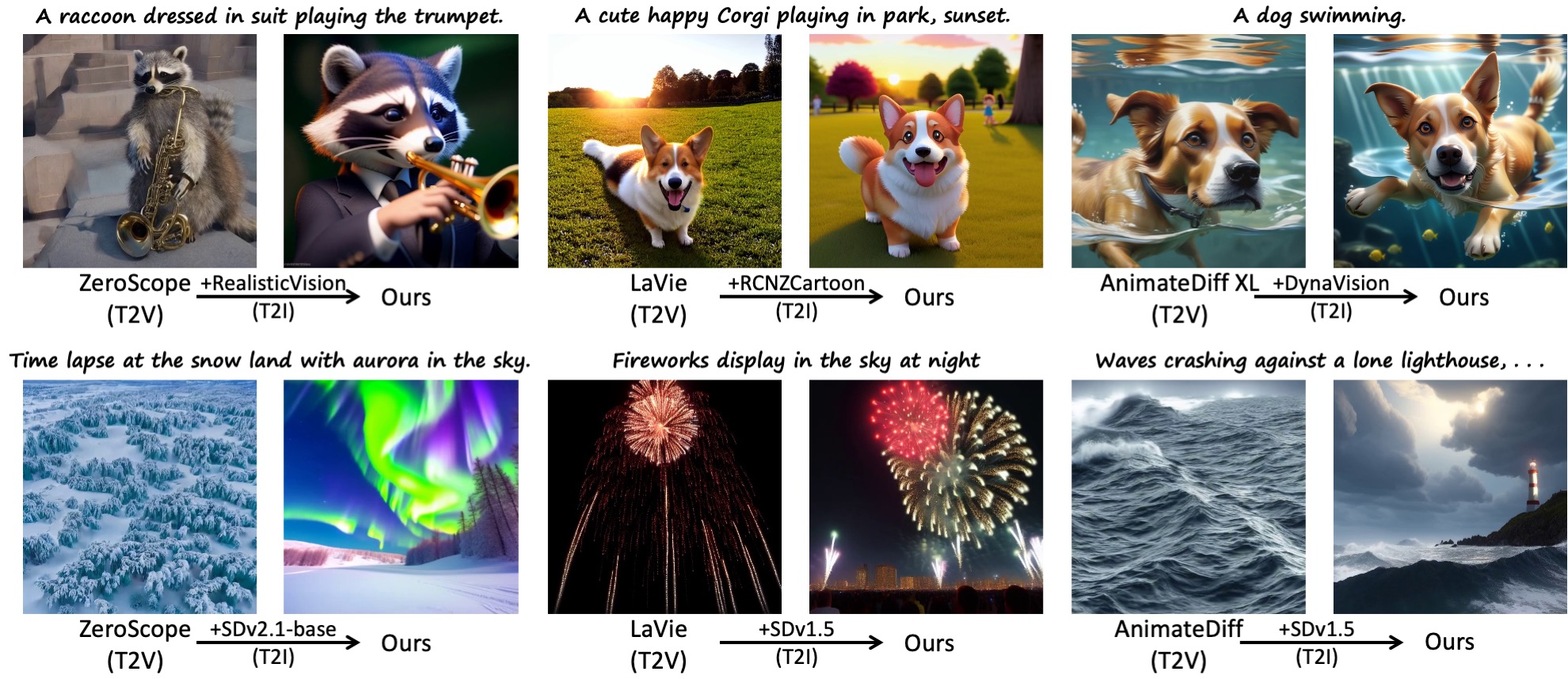}{figures/intro_teaser_jpg/}{1}{16}
    \vspace{-0.5em}
    \captionof{figure}
    {\textbf{Videos enhanced by VideoElevator}. 
    VideoElevator aims at elevating the quality of videos generated by existing text-to-video models (\eg, ZeroScope) with text-to-image diffusion models (\eg, RealisticVision). It is \textit{training-free} and \textit{plug-and-play} to support cooperation of various text-to-video and text-to-image diffusion models. \textit{Best viewed with Acrobat Reader. Click images to play the videos.}}
    \vspace{-0.5em}
\end{center}%

\begin{abstract}
Text-to-image diffusion models (T2I) have demonstrated unprecedented capabilities in creating realistic and aesthetic images.
On the contrary, text-to-video diffusion models (T2V) still lag far behind in frame quality and text alignment, owing to insufficient quality and quantity of training videos.
In this paper, we introduce VideoElevator, a \textit{training-free} and \textit{plug-and-play} method, which elevates the performance of T2V using superior capabilities of T2I.
Different from conventional T2V sampling (\ie, temporal and spatial modeling), VideoElevator explicitly decomposes each sampling step into \textit{temporal motion refining} and \textit{spatial quality elevating}.
Specifically, temporal motion refining uses encapsulated T2V to enhance temporal consistency, followed by inverting to the noise distribution required by T2I.
Then, spatial quality elevating harnesses inflated T2I to directly predict less noisy latent, adding more photo-realistic details.
We have conducted experiments in extensive prompts under the combination of various T2V and T2I.
The results show that VideoElevator not only improves the performance of T2V baselines with foundational T2I, but also facilitates stylistic video synthesis with personalized T2I.
Our code is available at \href{https://github.com/YBYBZhang/VideoElevator}{https://github.com/YBYBZhang/VideoElevator}
\end{abstract}
\section{Introduction}
\begin{figure}[ht]
   \begin{center}
   \includegraphics[width=.99\linewidth]{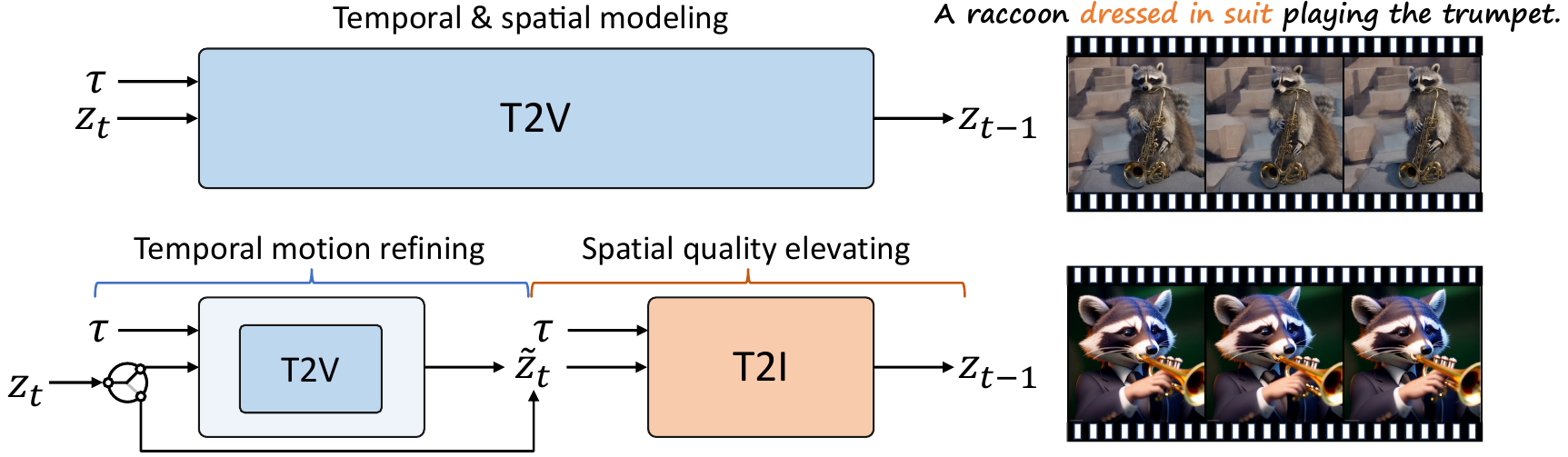}
   \end{center}
   \caption{\textbf{VideoElevator for improved text-to-video generation.}
   \textbf{Top: }Taking text $\tau$ as input, conventional T2V performs both temporal and spatial modeling and accumulates low-quality contents throughout sampling chain.
   \textbf{Bottom: }VideoElevator explicitly decompose each step into \textit{temporal motion refining} and \textit{spatial quality elevating}, where the former encapsulates T2V to enhance temporal consistency and the latter harnesses T2I to provide more faithful details, \eg, \textcolor{orange}{dressed in suit}.
   Empirically, applying T2V in several timesteps is sufficient to ensure temporal consistency.
   } 
    \label{fig:intro}
    \vspace{-1em}
\end{figure}

Diffusion models~\cite{sohl2015deep,ho2020ddpm,song2021score} have facilitated the rapid development of generative modeling, and demonstrated tremendous success in multiple modalities~\cite{poole2022dreamfusion,ramesh2022hierarchical,nichol2022glide,ho2022imagenvideo,ho2022video,singer2022make,blattmann2023align}, especially in image and video synthesis.
Nonetheless, exceptional generative capabilities primarily depend on an abundance of high-quality datasets.
For example, text-to-image diffusion models (T2I) requires billions of highly aesthetic images to achieve desirable control and quality~\cite{ramesh2022hierarchical,saharia2022photorealistic,rombach2022high,podell2023sdxl}.
However, owing to the difficulty in collection, the publicly available video datasets~\cite{bain2021frozen} are far behind in both quantity and quality (\ie, million-level scale and low-quality contents), which greatly limits the text-to-video diffusion models (T2V)~\cite{guo2023animatediff,wang2023modelscope,wang2023lavie,zhang2023show1} in prompt fidelity and frame quality.

Recent studies~\cite{ho2022video,ho2022imagenvideo,wang2023modelscope,guo2023animatediff,wang2023lavie,zhang2023show1} promisingly enhance video generation performance through the advances in T2I.
They either jointly train a T2V using video and image datasets~\cite{ho2022video,ho2022imagenvideo}, or initialize it with a pre-trained T2I~\cite{wang2023modelscope,guo2023animatediff,wang2023lavie,zhang2023show1}.
With the spatial modeling capabilities of T2I, T2V can concentrate more on the learning of temporal dynamics.
However, as training progresses, T2V inevitably will be biased towards low visual quality caused by the training videos~\cite{bain2021frozen}, such as lower resolutions, blurriness, and inconsistent descriptions.
Even when integrating with higher-quality personalized T2I, AnimateDiff~\cite{guo2023animatediff} cannot rectify the bias towards low visual quality in temporal layers.
As a result, synthesized videos still suffer from the frame quality degradation issue in comparison to T2I generated images, and sometimes contain visible flickers.

Considering that low-quality contents accumulate throughout T2V sampling chain, we investigate to rectify it with high-quality T2I at each step.
As illustrated in Fig.~\ref{fig:intro}, we explicitly decompose each sampling step into two components: (i) temporal motion refining \textit{and} (ii) spatial quality elevating.
Temporal motion refining encapsulates T2V to produce more temporally-consistent video latent and then inverts the latent to current timestep, so that spatial quality elevating can directly leverage T2I to add realistic details.
Compared to conventional T2V sampling, the decomposed sampling potentially raises the bar of synthesized video quality.

We present VideoElevator, a \textit{training-free} and \textit{plug-and-play} approach, as a way of improving video generation performance with decomposed sampling.
Given video latent $\vz_{t}$ at timestep $t$, \textit{temporal motion refining} encapsulates T2V with a temporal low-pass frequency filter to improve consistency, followed by T2V-based SDEdit~\cite{meng2021sdedit} to portray natural motion.
To obtain noise latent $\tilde{\vz}_{t}$ required by T2I, it deterministically inverts~\cite{song2021denoising} video latent to preserve motion as much as possible.
After that, \textit{spatial quality elevating} leverages inflated T2I to directly convert $\tilde{\vz}_{t}$ to less noisy $\vz_{t-1}$, where the self-attention in T2I is extended into cross-frame attention for appearance consistency.
Empirically, applying temporal motion refining in several timesteps is sufficient to ensure temporal consistency, so we omit it in other timesteps for efficiency.
To ensure interaction between various T2V and T2I, VideoElevator projects all noise latents to clean latents before being fed into another model.
Benefiting from it, VideoElevator supports the combination of various T2V and T2I, as long as their clean latent distributions are shared (\ie, same autoencoders).

Evaluating on extensive video prompts, VideoElevator not only greatly improves the generation quality of T2V with foundational T2I, but also facilitates creative video synthesis with versatile personalized T2I.
Firstly, it visibly enhances a wide range of T2V with Stable Diffusion V1.5 or V2.1-base~\cite{rombach2022high}, achieving higher frame quality and prompt consistency than their baselines.
Secondly, it enables T2V to replicate the diverse styles of personalized T2I more faithfully than AnimateDiff~\cite{guo2023animatediff}, while supporting T2I with significant parameter shifts, \eg, DPO-enhanced T2I~\cite{wallace2023diffusion}.

To summarize, our key contributions are as three-fold:
\begin{itemize}
    \item We introduce VideoElevator, a \textit{training-free} and \textit{plug-and-play} method, which enhances the quality of synthesized videos with versatile text-to-image diffusion models.
    \item To enable cooperation between various T2V and T2I, we present two novel components \textit{temporal motion refining} and \textit{spatial quality elevating}, where the former applies encapsulated T2V to improve temporal consistency and the latter harnesses inflated T2I to provide high-quality details.
    \item The experiments show that VideoElevator significantly improves the performance of T2V baselines by leveraging versatile T2I, in terms of frame quality, prompt consistency, and aesthetic styles.
\end{itemize}

\section{Related Work}
\subsection{Text-to-image diffusion models}
\noindent \textbf{Foundational text-to-image diffusion models.}
Diffusion models have achieved unprecedented breakthroughs in image creation and editing tasks~\cite{ramesh2022hierarchical,nichol2022glide,saharia2022photorealistic,rombach2022high,podell2023sdxl,zhang2023adding,balaji2022ediffi}, significantly surpassing the performance of previous models based on GANs~\cite{zhang2017stackgan,kang2023gigagan,sauer2023stylegan} and auto-regressive models~\cite{ramesh2021zero,chang2023muse,yu2022scaling}.
Benefiting from large-scale exceptional training data, foundational text-to-image diffusion models can producing images with photo-realistic quality and textual mastery.
To reduce computational complexity, Latent Diffusion Model (LDM)~\cite{rombach2022high} applies diffusion process in the latent space of auto-encoders.
Stable Diffusion (SD) is the most popular open-sourced instantiation of LDM, and facilitates a surge of recent advances in downstream tasks~\cite{zhang2023adding,gal2022textualinversion,poole2022dreamfusion,ma2023magicstick,huang2023dreamcontrol,lin2023improving,ni2023ref,lv2024place}.
By increasing parameters and improving training strategies of SD, Stable Diffusion XL (SDXL)~\cite{podell2023sdxl} obtains drastically better performance than the original Stable Diffusion.

\noindent \textbf{Personalized text-to-image diffusion models.}
Adapting from foundational text-to-image models, personalized text-to-image models~\cite{gal2022textualinversion,ruiz2022dreambooth,kumari2022multi,wei2023elite,cai2023decoupled,wallace2023diffusion,kim2023architectural} can satisfy a wide range of user requirements, \eg, various aesthetic styles and human preference alignment.
In particular, DreamBooth~\cite{ruiz2022dreambooth} and LoRA~\cite{hu2021lora} enable users to efficiently finetune foundation models on their customized small datasets, thereby promoting the release of various stylistically personalized models.
Diffusion-DPO~\cite{wallace2023diffusion} refines on large-scale preference benchmarks and aligns Stable Diffusion better with human preference.
Among them, DreamBooth and LoRA only slightly changes the parameters of foundation models, while Diffusion-DPO and Distilled-diffusion greatly alter their parameters.

\subsection{Text-to-video diffusion models}
\noindent \textbf{Direct text-to-video diffusion models.}
Most text-to-video diffusion models~\cite{chen2023videocrafter1,wang2023videocomposer,zhou2022magicvideo,he2022latent,wang2023modelscope,luo2023videofusion,wang2023lavie,zhang2023show1,esser2023structure,ma2023follow,khachatryan2023text2video,zhang2023controlvideo,wu2023tune,ge2023preserve,liang2023movideo,bar2024lumiere} are derived from text-to-image diffusion models, often incorporating additional temporal layers. 
VDM~\cite{ho2022video} and Image-Video~\cite{ho2022imagenvideo} jointly train models from the scratch using large-scale video-text and image-text pairs.
Based on pre-trained Stable Diffusion, LaVie~\cite{wang2023lavie} and ZeroScope~\cite{wang2023modelscope} finetune the whole model during training.
Differently, VideoLDM~\cite{blattmann2023align} and AnimateDiff~\cite{guo2023animatediff} only finetune the additional temporal layers, enabling them to be plug-and-play with personalized image models.
We note that Stable Diffusion based video models frozen the pre-trained autoencoders and thus share the same clean latent distribution. 
Compared to AnimateDiff~\cite{guo2023animatediff}, our VideoElevator produces higher-quality videos and supports a wider range of personalized image models, \eg, Diffusion-DPO~\cite{wallace2023diffusion}.

\noindent \textbf{Factorized text-to-video diffusion models.}
Due to the low-quality contents in training videos~\cite{chen2023videocrafter1,wang2023videocomposer,zhou2022magicvideo,he2022latent,wang2023modelscope,luo2023videofusion,wang2023lavie,zhang2023show1}, direct text-to-video diffusion models generates videos of much lower quality and fidelity than the image counterparts~\cite{rombach2022high,saharia2022photorealistic}.
To alleviate this issue, factorized text-to-video diffusion models~\cite{li2023generative,li2023videogen,singer2022make,girdhar2023emu,blattmann2023stable,xing2023dynamicrafter,zhang2023i2vgen,zeng2023make,dai2023fine} enhance the performance from two perspectives: (i) factorize the generation into text-to-image and image-to-video synthesis and (ii)
internally re-collect a large amount of high-quality video data.
Make-A-Video~\cite{singer2022make} and I2VGen-XL~\cite{zhang2023i2vgen} replace the text embedding with more informative image embedding as input condition.
EMU-Video~\cite{girdhar2023emu} and Make-Pixels-Dance~\cite{zeng2023make} produce the first frame with the state-of-the-art image models, following by concatenating them to generate high-quality videos.
In contrast, our VideoElvator requires no extra training and faithfully integrates the superior capabilities of text-to-image models into sampling chain.

\section{Preliminary}
\noindent \textbf{Latent diffusion model.}
Instead of applying diffusion process in image space, latent diffusion model (LDM)~\cite{rombach2022high} performs this in lower-dimensional latent space.
LDM mainly consists of pre-trained autoencoder and U-Net architecture~\cite{ronneberger2015unet}.
The autoencoder uses an encoder $\encoder$ to compress an image $\vx$ into latent code $\vz=\encoder(\vx)$ and a decoder $\decoder$ to reconstruct it.
The U-Net is trained to learn the clean latent distribution $\vz_0 \sim p_{data}(\vz_0)$ within the DDPM framework~\cite{song2021score,sohl2015deep,ho2020ddpm}.
In specific, LDM performs a forward diffusion process by adding Gaussian noise to compute $\vz_t$:
\begin{align}
    q(\vz_t|\vz_{0}) = \mathcal{N}(\vz_t; \sqrt{\alpha_t}\vz_{0}, \sqrt{1-\alpha_t} \mathbf{I}),
    \label{eq:forward}
\end{align}
where $T$ is the number of diffusion timesteps and $\{\alpha_{t}\}_{t=1}^{T}$ control the noise schedules.
With the U-Net architecture, LDM learns to reverse the above diffusion process through predicting less noisy $\vz_{t-1}$:
\begin{equation}
    p_\theta(\vz_{t-1}|\vz_t) = \mathcal{N}(\vz_{t-1};\mu_\theta(\vz_t,t),\Sigma_\theta(\vz_t,t)),
    \label{eq:backward}
\end{equation}
$\mu_\theta$ and $\Sigma_\theta$ are the mean and variance parameterized with learnable $\theta$.
Using LDM (\eg, Stable Diffusion) as base model, T2V and personalized T2I usually freeze the autoencoder during finetuning, so they share clean latent distribution.

\noindent \textbf{DDIM sampling and inversion.}
According to Eqn.~\ref{eq:forward}, one can directly predict the clean latent $\vz_{t\rightarrow 0}$ with the noise latent $\vz_{t}$:
\begin{align}
    \vz_{t\rightarrow 0} = \frac{\vz_t - \sqrt{1 - \alpha_t} \epsilon_\theta(\vz_t, t)}{\sqrt{\alpha_t}},
    \label{eq:project}
\end{align}
where $\epsilon_\theta(\cdot, \cdot)$ denotes ``$\epsilon$-prediction'' diffusion model.
When producing new samples from a random noise $\vz_{T} \sim \mathcal{N}(\mathbf{0}, \mathbf{I})$, DDIM sampling~\cite{song2021denoising} denoises $\vz_{t}$ to $\vz_{t-1}$ of previous timestep:
\begin{align}
    \vz_{t-1} & = \sqrt{\alpha_{t-1}} \vz_{t\rightarrow 0} + \sqrt{1 - \alpha_{t-1}} \cdot \epsilon_\theta(\vz_t, t),
    \label{eq:ddim_samp}
\end{align}

Given a clean latent $\vz_0$, DDIM inversion~\cite{song2021denoising} deterministically inverts it into noise latent $\vz_T$:
\begin{align}
    \vz_{t} & = \sqrt{\alpha_{t}} \vz_{(t-1)\rightarrow 0} + \sqrt{1 - \alpha_{t}} \cdot \epsilon_\theta(\vz_{t-1}, t),
    \label{eq:ddim_inv}
\end{align}
The inverted $\vz_T$ can reconstruct $\vz_0$ by iterating Eqn.~\ref{eq:ddim_samp}.
Thanks to its capability in keeping structure, DDIM inversion has been widely used in image and video editing tasks~\cite{hertz2022prompt,wu2023tune}.

\noindent \textbf{Noise schedules and latent distributions.}
Noise schedules are parameterized by $\{\alpha_{t}\}_{t=1}^{T}$ and define the noise scales at different timesteps~\cite{ho2020ddpm,song2021score}.
In general, T2V and T2I adopt different noise schedules during training, \ie, $\{\alpha_{t}^V\}_{t=1}^{T} \neq \{\alpha_{t}^I\}_{t=1}^{T}$.
Therefore, their noise latent codes belong to different noise distributions, and cannot be used as input for each other.

\section{VideoElevator}
\begin{figure}[t]
   \begin{center}
   \includegraphics[width=.99\linewidth]{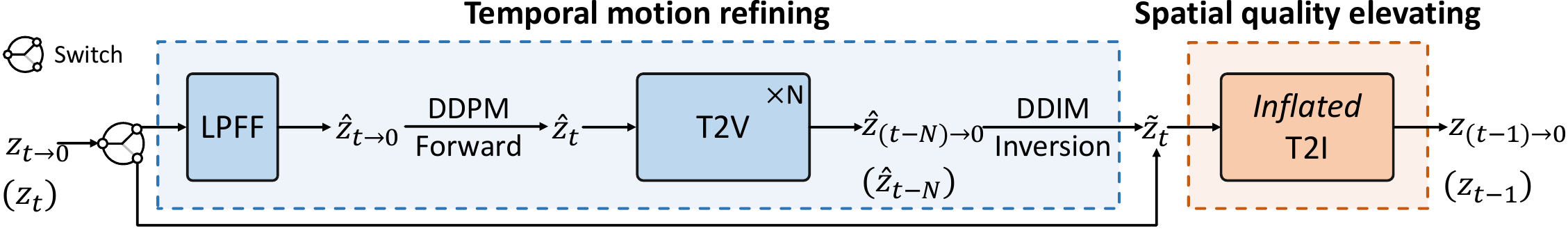}
   \end{center}
   \caption{\textbf{Overview of VideoElevator.}
   VideoElevator explicitly decomposes each sampling step into temporal motion refining and spatial quality elevating.
   \textbf{Temporal motion refining} uses a Low-Pass Frequency Filter (LPFF) to reduce flickers and T2V-based SDEdit~\cite{meng2021sdedit} to add fine-grained motion, and then inverts the latent to $\tilde{\vz}_{t}$ with DDIM inversion~\cite{song2021denoising}.
   \textbf{Spatial quality elevating} harnesses \textit{inflated} T2I to directly transition $\tilde{\vz}_{t}$ to $\vz_{t-1}$, where self-attention of T2I is inflated into cross-frame attention.
   To ensure interaction between T2V and T2I, noise latents are equally projected to clean latents with Eqn.~\ref{eq:project}, \eg, $\vz_{t}$ to $\vz_{t\rightarrow 0}$.
   } 
    \label{fig:method}
\end{figure}
We introduce VideoElevator to enhance the performance of T2V by integrating T2I into sampling chain, which consists of temporal motion refining (in Sec.~\ref{sec:temporal}) and spatial quality elevating (in Sec.~\ref{sec:spatial}).
Specifically, VideoElevator reformulates the sampling step at timestep $t$ as follows:
taking noise latent $\vz_{t}$ as input, temporal motion refining encapsulates T2V to enhance temporal consistency and inverts noise latent to $\tilde{\vz}_{t}$, so that spatial quality elevating directly harnesses T2I to transition it to higher-quality $\vz_{t-1}$.

Since T2I and T2V are usually trained in different noise schedules, their noise latent at timestep $t$ may belong to different distributions and cannot be fed into another model.
To ensure their interaction, we first project the output $\vz_{t}$ of T2I and $\hat{\vz}_{t-N}$ of T2V into clean latent $\vz_{t\rightarrow 0}$ and $\hat{\vz}_{(t-N)\rightarrow 0}$ with Eqn.~\ref{eq:project}.
Then, we forward clean latent to the noise distributions required by T2V and T2I, but adopt DDPM forward and DDIM inversion respectively.
Benefiting from it, VideoElevator supports the combination of various T2V and T2I, as long as their clean latent distributions are shared (\ie, same autoencoder).

\subsection{Temporal motion refining}
\label{sec:temporal}
Temporal motion refining mainly involves in two aspects: (i) enhance temporal consistency of video latent $\vz_{t\rightarrow 0}$ using T2V generative priors, and (ii) convert it to the noise distribution required by T2I.
Firstly, the video decoded from $\vz_{t\rightarrow 0}$ lacks reasonable motion and contains visible flickers.
With the power of T2V generative priors~\cite{meng2021sdedit}, it is possible to generate a video with natural motion, but challenging to remove its visible flickers (refer to Fig.~\ref{fig:ab_filter} (a)).
Therefore, before using T2V generative priors, we first transform $\vz_{t\rightarrow 0}$ into frequency domain to reduce its high-frequency flickers.
Secondly, to preserve the motion integrity in (ii), we deterministically invert video latent into the corresponding noise distribution of T2I.

Specifically, in Fig.~\ref{fig:method}, we first apply low-pass frequency filter $\texttt{LPFF}(\cdot)$ in $\vz_{t\rightarrow 0}$ along the \textit{temporal} dimension, which computes more stable $\hat{\vz}_{t\rightarrow 0}$:
\begin{align}
    \boldsymbol{\mathcal{F}}(\boldsymbol{\vz}_{t\rightarrow 0}) &= \texttt{FFT}_{Temp}(\boldsymbol{\vz}_{t\rightarrow 0}), \\
    \hat{\boldsymbol{\mathcal{F}}}(\boldsymbol{\vz}_{t\rightarrow 0}) &= \boldsymbol{\mathcal{F}}(\boldsymbol{\vz}_{t\rightarrow 0}) \odot \boldsymbol{\mathcal{G}}, \\
    \hat{\boldsymbol{\vz}}_{t\rightarrow 0} &= \texttt{IFFT}_{Temp}(\hat{\boldsymbol{\mathcal{F}}}(\boldsymbol{\vz}_{t\rightarrow 0})),
    \label{eq:low_pass}
\end{align}
where $\texttt{FFT}_{Temp}(\cdot)$ is the fast fourier transformation to map $\vz_{t\rightarrow 0}$ to the frequency domain, while $\texttt{IFFT}_{Temp}(\cdot)$ denotes inverse fast fourier transformation to map it back to the temporal domain.
$\mathcal{G}$ represents Gaussian low-pass filter mask to diminish high-frequency flickers.
Compared to \textit{spatial-temporal} \texttt{LPFF}~\cite{wu2023freeinit}, our \textit{temporal} \texttt{LPFF} not only effectively stabilizes the video, but also has less negative impact on spatial quality, \eg, more realistic details in Fig.~\ref{fig:ab_filter}.

Albeit there are less high-frequency flickers in $\hat{\vz}_{t\rightarrow 0}$, it is insufficient to achieve vivid and fine-grained motion without T2V generative priors.
Inspired by~\cite{meng2021sdedit}, we adopt T2V-based SDEdit to portray natural motion to $\hat{\vz}_{t\rightarrow 0}$.
In particular, with T2V noise schedule $\{\alpha_{t}^V\}_{t=1}^{T}$, we perform DDPM forward process in $\hat{\vz}_{t\rightarrow 0}$ to calculate $\hat{\vz}_{t}$ (with Eqn.~\ref{eq:forward}), and then iterates the T2V denoising process $N$ times to achieve desirable motion (with Eqn.~\ref{eq:ddim_samp}), \ie, $\hat{\vz}_{t-N}$ or $\hat{\vz}_{(t-N)\rightarrow{0}}$.
Notably, when $N$ is very small (\eg, $N=1$), the synthesized video only contains coarse-grained motion, so we set $N$ to $8\sim 10$ to add fine-grained one (refer to Appendix~\textcolor{red}{B}).

Finally, when inverting $\hat{\vz}_{(t-N)\rightarrow{0}}$ into the input noise distribution of T2I, we deterministically forward it to keep its motion.
Using the unconditional guidance~\cite{ho2022classifier} (\ie, null condition $\emptyset$), clean latent $\hat{\vz}_{(t-N)\rightarrow{0}}$ is inverted to noise latent distribution of T2I at timestep $t$:
\begin{align}
    \tilde{\vz}_{t} = \texttt{Inversion}(\epsilon_\phi^I; \hat{\vz}_{(t-N)\rightarrow{0}}, t)
\end{align}
where $\epsilon_\phi^I(\cdot, \cdot, \cdot)$ represents inflated T2I.
We use $\texttt{Inversion}(\cdot; \cdot, \cdot)$ to denote a diffusion-based inversion algorithm, and choose the DDIM inversion~\cite{song2021denoising} by default.
In contrast, DDPM-based strategies potentially impair the motion integrity of synthesized videos, \eg, leading to all frames being identical or inconsistent in Fig.~\ref{fig:ab_inversion}.

Empirically, applying temporal motion refining in just a few timesteps (\ie, $4\sim 5$ steps) can ensure temporal consistency (refer to Appendix \textcolor{red}{B}).
To improve sampling efficiency, we perform both temporal motion refining and spatial quality elevating in selected timesteps, while performing T2I denoising only in other timesteps.

\subsection{Spatial quality elevating}
\label{sec:spatial}

Given stablized latent $\tilde{\vz}_{t}$ from temporal motion refining, spatial quality elevating leverages T2I to directly add high-quality details.
However, individually denoising all frames with conventional T2I will lead to visible inconsistency in appearance.
Motivated by previous works~\cite{zhang2023controlvideo,wu2023tune,khachatryan2023text2video}, we inflate T2I along the temporal axis so that all frames share the same content.

Particularly, we extend the U-Net of T2I along the temporal dimension, including convolution and self-attention layers.
The 2D convolution layers are converted to 3D counterpart by replacing $3 \times 3$ kernels with $1\times 3 \times 3$ kernels.
The self-attention layers are extended to first-only cross-frame attention layers by adding inter-frame interaction.
For first-only cross-frame interaction, frame latents $\tilde{\vz}_{t}=\{\tilde{\vz}_{t}[i]\}_{i=0}^{F-1}$ ($F$ is the number of frames) are updated as:
{
\begin{align}
    \mathrm{Attention}(\mQ,\mK,\mV) =\mathrm{Softmax}(\frac{\mQ \mK^T}{\sqrt{d}}) \cdot \mV, 
    \label{eq:cross_frame}
\end{align}
}
where $\mQ =\mW^Q \tilde{\vz}_{t}[i]$, $\mK=\mW^K \tilde{\vz}_{t}[0]$,  and $\mV=\mW^V \tilde{\vz}_{t}[0]$.

Afterwards, we directly utilize \textit{inflated} T2I to add photo-realistic details to $\tilde{\vz}_{t}$, transitioning it to less noisy latent $\vz_{t-1}$ at timestep $t-1$:
\begin{align}
    \vz_{t-1} = \sqrt{\alpha_{t-1}^I} \cdot \tilde{\vz}_{t\rightarrow 0} + \sqrt{1 - \alpha_{t-1}^I} \cdot \epsilon_\phi^I(\tilde{\vz}_{t}, y, t) + \sigma_t^I \epsilon_t.
\end{align}
where $\epsilon_{t}$ is random Gaussian noise and $\sigma_{t}^I$ controls its scale.

At each sampling step, spatial quality elevating employs inflated T2I to provide realistic details to video latent.
Compared to T2V baselines, VideoElevator produces videos whose frames are closer to T2I generated images, from the perspectives of text alignment, quality, and aesthetic styles.

\begin{figure}[t!]
   \begin{center}
   \includegraphics[width=.99\linewidth]{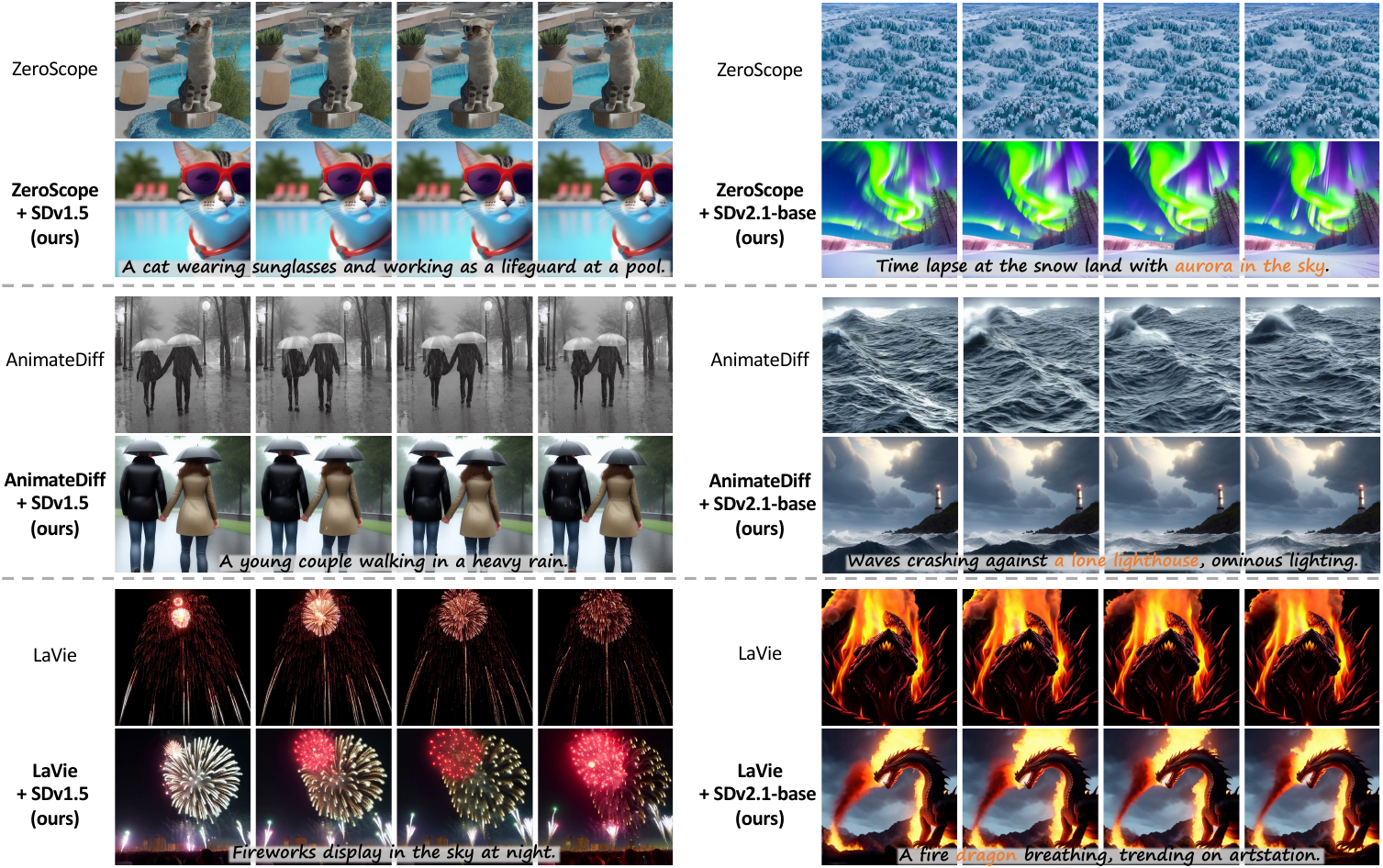}
   \end{center}
    \vspace{-1em}
   \caption{\textbf{Qualitative results enhanced with foundational T2I.}
   As one can see, VideoElevator manages to enhance the performance of T2V baselines with StableDiffusion V1.5 or V2.1-base, in terms of frame quality and text alignment.
   For frame quality, the videos enhanced by VideoElevator contain more details than original videos.
   For text alignment, VideoElevator also produces videos that adhere better to prompts, where inconsistent parts of baselines are colored in \textbf{\textcolor{orange}{orange}}.
   \textit{Please watch videos in website for better view.}
   } 
    \label{fig:main_base}
    \vspace{-1em}
\end{figure}
\begin{figure}[ht]
   \begin{center}
   \includegraphics[width=.99\linewidth]{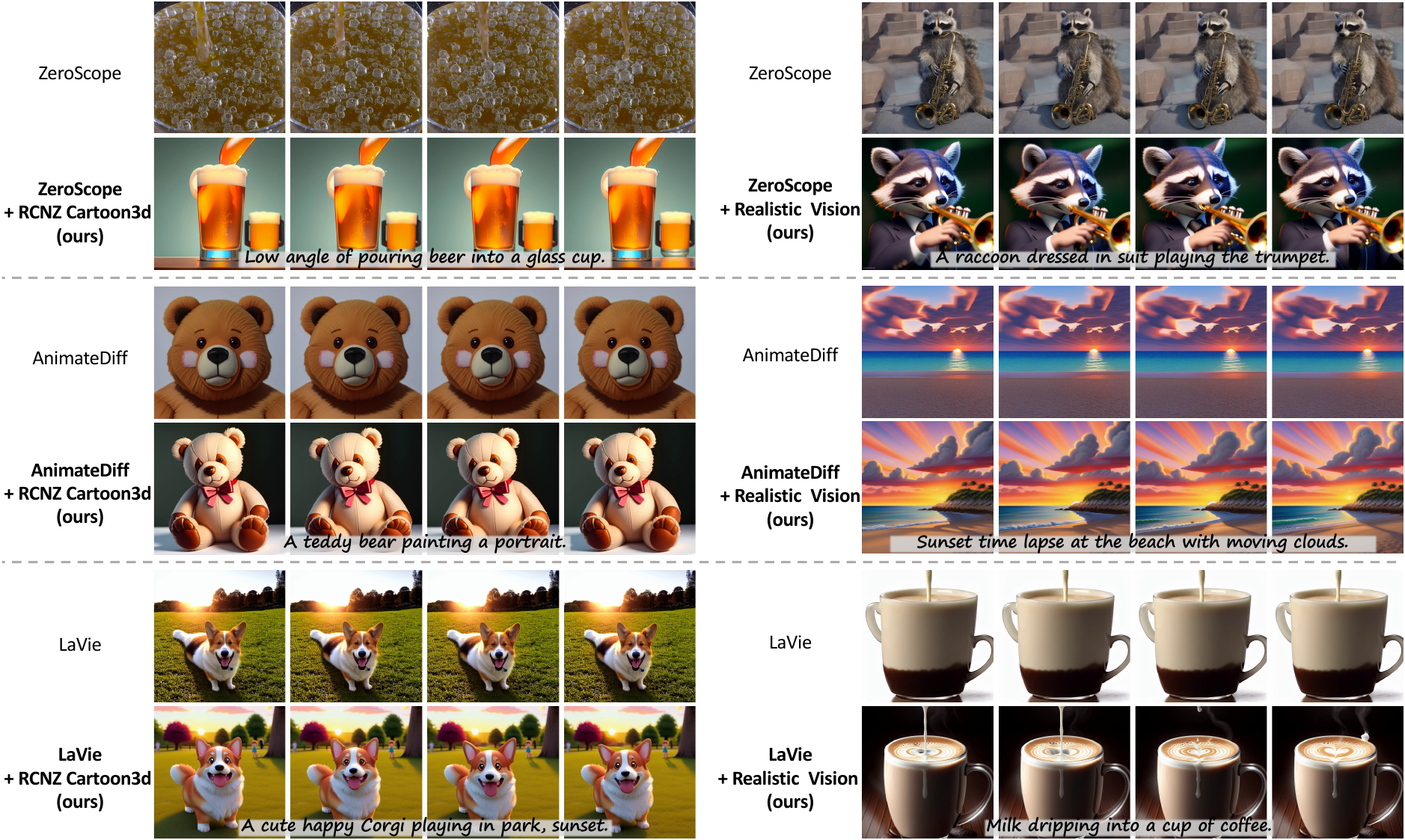}
   \end{center}
    \vspace{-1em}
   \caption{\textbf{Qualitative results enhanced with personalized T2I.}
   With the power of personalized T2I, VideoElevator enables ZeroScope and LaVie to produce various styles of high-quality videos.
   Compared to personalized AnimateDiff, VideoElevator captures more faithful styles and photo-realistic details from personalized T2I, \eg, \texttt{sunset time lapse at the beach}.
   \textit{Please watch videos in website for better view.}
   } 
    \label{fig:main_lora}
    \vspace{-1em}
\end{figure}
\section{Experiments}
\subsection{Experimental settings}
\noindent \textbf{Baselines.}
We choose three popular text-to-video diffusion models as our baselines and enhance them with versatile text-to-image models:
(i) LaVie~\cite{wang2023lavie} that is initialized with Stable Diffusion V1.4;
(ii) AnimateDiff~\cite{guo2023animatediff} that is inflated from Stable Diffusion V1.5;
(iii) ZeroScope~\cite{wang2023modelscope} that is initialized with Stable Diffusion V2.1-base.
Among them, LaVie and ZeroScope finetune the whole U-Net.
AnimateDiff only finetunes the temporal layers and support to integrate the personalized T2I based on DreamBooth~\cite{ruiz2022dreambooth} or LoRA~\cite{hu2021lora}.
When enhancing them with personalized T2I, VideoElevator uses the base version of LaVie and ZeroScope, and personalized version of AnimateDiff.

\noindent \textbf{Evaluation benchmarks.}
We evaluate VideoElevator and other baselines in two benchmarks:
(i) VBench~\cite{huang2023vbench} dataset that involves in a variety of content categories and contains $800$ prompts;
(ii) VideoCreation dataset, which unifies creative prompts datasets of Make-A-Video~\cite{singer2022make} and VideoLDM~\cite{blattmann2023align} and consists of $100$ prompts in total.
We use VBench dataset to evaluate T2V enhanced by foundational T2I, and choose VideoCreation dataset for T2V improved by personalized T2V as well as human evaluation.

\noindent \textbf{Automated metrics.}
We utilize five metrics to comprehensively evaluate the performance of T2V:
(i) Frame consistency (FC)~\cite{esser2023structure,zhang2023controlvideo} that calculates the mean CLIP score between each pair of adjacent frames;
(ii) Prompt consistency (PC) that computes the mean CLIP score between text prompt and all frames;
(iii) Frame quality (FQ) based on CLIP-IQA~\cite{wang2023exploring} to quantify the perceived quality, \eg, distortions;
(iv) Aesthetic score (AS)~\cite{huang2023vbench,LAIONaes} that assesses artistic and beauty value of each frame using the LAION aesthetic predictor;
(v) Domain similarity (DS)~\cite{guo2023animatediff} that computes the mean CLIP score between all video frames and a reference image.
We implement all above metrics based on CLIP ViT-L/14~\cite{radford2021learning}.

\subsection{Comparisons with T2V baselines}
We verify the effectiveness of VideoElevator under two types of T2I, including foundational and personalized ones (refer to Appendix~\textcolor{red}{A} for details).

\noindent \textbf{Qualitative comparison.}
Fig.~\ref{fig:main_base} visualizes the results of three T2V enhanced with foundational T2I.
As one can see, VideoElevator visibly improves the frame quality and prompt fidelity of synthesized videos with either Stable Diffusion V1.5 or V2.1-base.
In row $1$ and $2$ of Fig.~\ref{fig:main_base}, VideoElevator makes the synthesized videos align better with text prompts than ZeroScope and AnimateDiff, \eg, lacking of \textcolor{orange}{aurora in the sky} and \textcolor{orange}{a lone lighthouse}.
Additionally, in row $3$, the videos of VideoElevator have more photo-realistic details than that of LaVie, \eg, more colorful fireworks.

Fig.~\ref{fig:main_lora} presents the generated videos empowered with a variety of personalized T2I.
From Fig.~\ref{fig:main_lora}, VideoElevator supports to inherit diverse styles from personalized T2I more faithfully than the alternative competitors.
In contrast, personalized AnimateDiff~\cite{radford2021learning} captures less-fidelity styles than ours, since its temporal layers suffer from low-quality contents in training videos.
ZeroScope~\cite{wang2023modelscope} and LaVie~\cite{wang2023lavie} are not compatible with personalized T2I, as they largely modify the parameters and feature space of initialized T2I.

\begin{table}[t]
\caption{
\textbf{Quantitative results} using foundational T2I.
We evaluate three T2V baselines and their enhanced versions on VBench benchmark.
The results of foundational T2I are provided as upper bound.
With the help of VideoElevator, both SDv1.5 and SDv2.1-base are successfully integrated into T2V baselines to improve their performance.
The best and second best results are \textbf{bolded} and \underline{underlined}.
}
\vspace{0.5em}
\centering
\scalebox{0.9}{
\begin{tabular}{lcccc}
\toprule
Method & Frame cons. & Prompt cons. &Aesthetic score & Frame quality`
\\
\midrule
\textcolor{gray}{SDv1.5} &\textcolor{gray}{$\backslash$} &\textcolor{gray}{0.264} &\textcolor{gray}{0.633} &\textcolor{gray}{0.758} \\
\textcolor{gray}{SDv2.1-base} &\textcolor{gray}{$\backslash$} &\textcolor{gray}{0.263} &\textcolor{gray}{0.646} &\textcolor{gray}{0.714} \\
\midrule
ZeroScope &0.983 &0.245 &0.561 &0.517 \\
ZeroScope+SDv1.5 (\textbf{Ours}) &\underline{0.987} &\textbf{0.252} &\underline{0.603} &\underline{0.576} \\
ZeroScope+SDv2.1-base (\textbf{Ours}) &\textbf{0.989} &\underline{0.248} &\textbf{0.618} &\textbf{0.593} \\
\midrule
AnimateDiff &0.983 &0.248 &0.582 &0.561 \\
AnimateDiff+SDv1.5 (\textbf{Ours}) &\underline{0.984} &\textbf{0.252} &\underline{0.610} &\underline{0.617} \\
AnimateDiff+SDv2.1-base (\textbf{Ours}) &\textbf{0.985} &\textbf{0.252} &\textbf{0.621} &\textbf{0.621} \\
\midrule
LaVie &0.991 &0.255 &0.607 &0.681 \\
LaVie+SDv1.5 (\textbf{Ours}) &\textbf{0.993} &\textbf{0.259} &\underline{0.627} &\textbf{0.706} \\
LaVie+SDv2.1-base (\textbf{Ours}) &\underline{0.992} &\underline{0.256} &\textbf{0.632} &\underline{0.702} \\
\bottomrule
\end{tabular}
}
\label{tab:main_base}
\vspace{-0.5em}
\end{table}
\begin{table}[t]
\caption{
    \textbf{User preference study.}
    The numbers denote the percentage of raters who favor the videos synthesized by VideoElevator over other baselines.}
    \vspace{2mm}
    \centering
    \setlength{\tabcolsep}{4.5pt}
    \scalebox{0.99}{
    \begin{tabular}{lccc}
    \toprule
         {Method comparison} & {Temporal consistency} & {Text alignment} & {Frame  quality} \\
         \midrule
         {Ours vs. ZeroScope} & $65\%$ & $81\%$ & $88\%$ \\
         {Ours vs. AnimateDiff} & $62\%$ & $75\%$ &$86\%$\\
         {Ours vs. LaVie} & $61\%$ & $72\%$ &$81\%$\\
    \bottomrule
    \end{tabular}
    }
    \label{tab:main_user}
\end{table}
\begin{table}[ht]
\caption{
\textbf{Quantitative results} using personalized T2I.
We evaluate three T2V baselines and their enhanced versions on VideoCreation dataset.
Compared to baselines, VideoElevator significantly improves their frame quality and aesthetic score, while capturing more faithful styles.
We provide the details of personalized T2I in Appendix.
The best results are \textbf{bolded}.
}
\vspace{-0.5em}
\centering
\scalebox{0.85}{
\begin{tabular}{lccccc}
\toprule
Method & Frame cons. & Prompt cons. &Aesthetic score & Frame qual. & Domain cons.
\\
\midrule
\textcolor{gray}{SD-LoRA} &\textcolor{gray}{$\backslash$} &\textcolor{gray}{0.288} &\textcolor{gray}{0.702} &\textcolor{gray}{0.790} &\textcolor{gray}{1.000}\\
\midrule
ZeroScope &0.981 &0.276 &0.578 &0.544 &0.700 \\
ZeroScope+SD-LoRA (\textbf{Ours}) &\textbf{0.991} &\textbf{0.281} &\textbf{0.671} &\textbf{0.684} &\textbf{0.871} \\
\midrule
AnimateDiff &0.993 &0.275 &0.667 &0.670 &0.830 \\
AnimateDiff+SD-LoRA (\textbf{Ours}) &\textbf{0.993} &\textbf{0.280} &\textbf{0.677} &\textbf{0.689} &\textbf{0.873} \\
\midrule
LaVie &0.991 &0.281 &0.628 &0.728 &0.773 \\
LaVie+SD-LoRA (\textbf{Ours}) &\textbf{0.992} &\textbf{0.285} &\textbf{0.680} &\textbf{0.709} &\textbf{0.852} \\
\bottomrule
\end{tabular}
}
\label{tab:main_lora}
\end{table}
\begin{figure}[ht]
   \begin{center}
   \includegraphics[width=.99\linewidth]{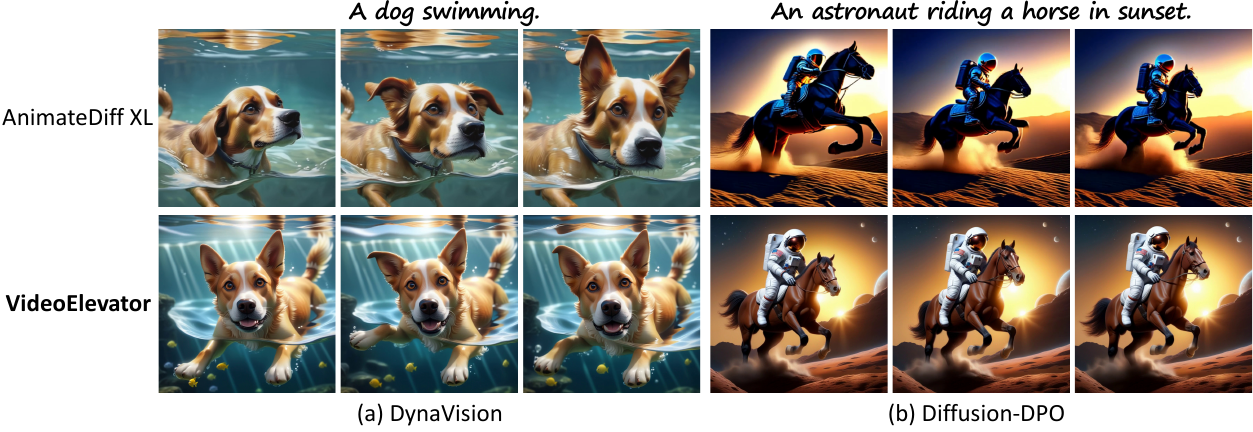}
    \vspace{-1em}
   \end{center}
   \caption{\textbf{Qualitative results using personalized SDXL.}
   We select AnimateDiff XL as T2V baseline and leverage two types of personalized SDXL to enhance it, including \textbf{(a)} DynaVision finetunes SDXL based on LoRA and slightly updates its parameters. \textbf{(b)} Diffusion-DPO largely updates the parameters of SDXL.
   } 
    \label{fig:main_sdxl}
    \vspace{-0.5em}
\end{figure}
\noindent \textbf{Quantitative comparison.}
Table~\ref{tab:main_base} shows the results of T2V integrated with foundational T2I.
In row $1$ and $2$ of Table~\ref{tab:main_base}, existing T2V lag far behind foundational T2I in prompt consistency, frame quality, and aesthetic score.
With the help of VideoElevator, all T2V baselines are significantly improved in aesthetic score and frame quality, with slightly better frame consistency.

Table~\ref{tab:main_lora} shows the results of T2V empowered by personalized T2I.
The results illustrate that VideoElevator effectively integrates high-quality T2I into existing T2V, especially in increasing their aesthetic scores.
In row $4$ and $5$ of Table~\ref{fig:main_lora}, VideoElevator is $0.043$ higher than personalized AnimateDiff in terms of domain consistency, \ie, capturing higher-fidelity styles, which is consistent to qualitative results.

\noindent \textbf{Human evaluation.}
To verify the effectiveness of VideoElevator, we further perform human evaluation in VideoCreation dataset~\cite{singer2022make,blattmann2023align} (including $100$ creative prompts).
The evaluation is conducted in three open-sourced pipelines, which are enhanced with Stable Diffusion V2.1-base.
We provide each rater a text prompt and two generated videos from different versions of a baseline (in random order).
Then, they are asked to select the better video for each of three perspectives: (i) temporal consistency, (ii) text alignment, and (iii) frame quality.
Each sample is evaluated by five raters, and we take a majority vote as the final selection.
Table~\ref{tab:main_user} summarizes the voting results of raters.
As one can see, the raters strongly favor the videos from enhanced baselines rather than their base version.
In specific, VideoElevator manages to boost their performance in terms of three measurements, especially in text alignment and frame quality.

\subsection{Results empowered by personalized SDXL}
We also verify the effectiveness of VideoElevator in personalized Stable Diffusion XL (SDXL).
We choose the SDXL version of AnimateDiff (AnimateDiff XL) as the baseline and improve it using LoRA-based and DPO-enhanced models.
The corresponding results are shown in Fig.~\ref{fig:main_sdxl}.
As illustrated in Fig.~\ref{fig:main_sdxl} (a), compared to AnimateDiffXL, VideoElevator produces videos that have more consistent styles with the LoRA-based SDXL, featuring realistic quality and rich details.
In Fig.~\ref{fig:main_sdxl} (b), when integrating Diffusion-DPO with large parameter shifts, VideoElevator also exhibits better compatibility than AnimateDiff XL, where the latter generates videos with lower frame quality and less details.

\begin{table}[t]
\caption{
    \textbf{Quantitative ablation on low-pass frequency filter (LPFF).}
    The results are evaluated in VideoCreation dataset under the combination of ZeroScope and personalized T2I.
    }
    \centering
    \setlength{\tabcolsep}{4.5pt}
    \scalebox{0.99}{
    \begin{tabular}{ccccccc}
      \toprule
      Spatial-LPFF &Temporal-LPFF &FC &PC &AS &FQ &DC\\
      \midrule
      \ding{56} &\ding{56} &0.982 &0.279 &0.670 &0.681 &0.870 \\
      \ding{56} &\ding{52} &\textbf{0.991} &\textbf{0.281} &\textbf{0.671} &\textbf{0.684} &\textbf{0.871}\\
      \ding{52} &\ding{52} &0.987 &0.279 &0.660 &0.668 &0.863 \\
    \bottomrule
    \end{tabular}
    }
    \label{tab:ab_filter}
\end{table}
\begin{figure}[ht]
   \begin{center}
   \includegraphics[width=.99\linewidth]{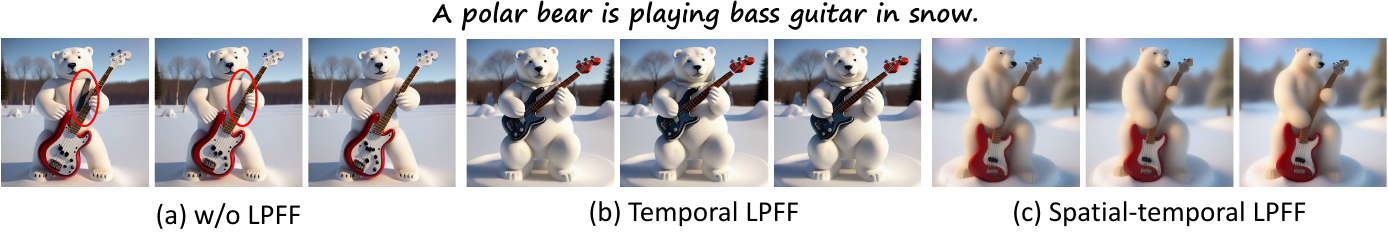}
   \end{center}
   \caption{\textbf{Qualitative ablation on low-pass frequency filter (LPFF).}
   \textbf{(a)} Not using LPFF leads to incoherence in appearance (highlighted in \textbf{\textcolor{red}{red circle}}). 
   \textbf{(b)} Using temporal LPFF visibly improves temporal consistency without hurting frame quality.
   \textbf{(c)} Adding both spatial and temporal LPFF makes video frames blurry.
   \textit{Please watch videos in website for better view.}
   } 
    \label{fig:ab_filter}
\end{figure}
\begin{table}[th]
\caption{
    \textbf{Quantitative ablation on inversion strategy.}
    The results are evaluated in VideoCreation dataset under the combination of ZeroScope and personalized T2I.
    }
    \centering
        \setlength{\tabcolsep}{4.5pt}
    \scalebox{0.99}{
        \begin{tabular}{lccccc}
              \toprule
              Inversion strategy &FC &PC &AS &FQ &DC\\
              \midrule
              \textcolor{gray}{Add same noise} &\textcolor{gray}{0.999} &\textcolor{gray}{0.278} &\textcolor{gray}{0.667} &\textcolor{gray}{0.683} &\textcolor{gray}{0.868}\\
              Add random noise &0.969 &0.278 &0.670 &0.683 &0.866 \\
              DDIM inversion &\textbf{0.991} &\textbf{0.281} &\textbf{0.671} &\textbf{0.684} &\textbf{0.871}\\
              \bottomrule
        \end{tabular}
    }
    \label{tab:ab_inversion}
\end{table}

\begin{table}[th]
\caption{
    \textbf{Quantitative ablation on sampling steps.}
    The results are evaluated in VideoCreation dataset.
    ZeroScope using $100$ steps and ZeroScope using $50$ steps perform similarly, both lagging behind our VideoElevator in a large margin.
    }
    \centering
    \setlength{\tabcolsep}{4.5pt}
    \scalebox{0.99}{
        \begin{tabular}{lcccccc}
          \toprule
          Method & Step number &FC &PC &AS &FQ &DC\\
          \midrule
          ZeroScope &50 &0.981 &0.276 &0.578 &0.544 &0.700 \\
          ZeroScope &100 &0.983 &0.275 &0.577 &0.553 &0.701 \\
          VideoElevator &100 &\textbf{0.991} &\textbf{0.281} &\textbf{0.671} &\textbf{0.684} &\textbf{0.871} \\
          \bottomrule
        \end{tabular}
    }
    \label{tab:ab_steps}
\end{table}

\begin{figure}[h!]
   \begin{center}
   \includegraphics[width=.99\linewidth]{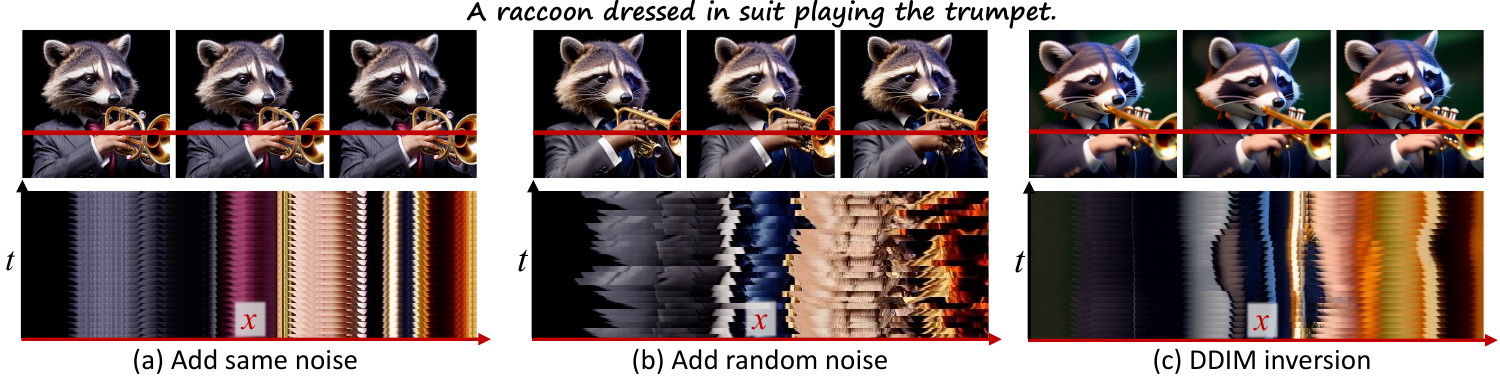}
   \end{center}
   \caption{\textbf{Quantitative ablation on inversion strategies.}
   We visualize synthesized frames and $x$-$t$ slice of pixels in \textbf{\textcolor{red}{red line}}.
   \textbf{(a)} Adding same noise for each frame makes all frames identical.
   \textbf{(b)} Adding random noise leads to noticeable discontinuity in $x$-$t$ slice.
   \textbf{(c)} Using DDIM inversion achieves more continuous $x$-$t$ slice across time.
   \textit{Please watch videos in website for better view.}
   } 
    \label{fig:ab_inversion}
\end{figure}
\subsection{Ablation studies}
We conduct more ablation experiments on the hyper-parameters of temporal motion refining, please refer to Appendix~\textcolor{red}{B} for details.

\noindent \textbf{Effect of low-pass frequency filter.}
Low-pass frequency filter (LPFF) aims to reduce the high-frequency flickers in video latents.
We probe its effect under three settings: (i) without LPFF, (ii) temporal LPFF, and (iii) spatial-temporal LPFF~\cite{wu2023freeinit}.
Their experimental results are shown in Fig.~\ref{fig:ab_filter} and Table~\ref{tab:ab_filter}.
In Fig.~\ref{fig:ab_filter} (a), w/o LPFF results in appearance incoherency, \eg, \textcolor{red}{disappearing black tape}.
In contrast, adding LPFF along temporal dimension noticeably improves frame consistency.
However, applying LPFF along both spatial and temporal dimensions visibly degrades frame quality and aesthetic score, making frames blurry and less detailed.

\noindent \textbf{Effect of inversion strategies.}
The inversion projects clean latent $\hat{\vz}_{(t-N)\rightarrow{0}}$ of T2V into noise latent $\tilde{\vz}_{t}$ of T2I at timestep $t$, aiming to inherit its natural motion.
We investigate three inversion strategies in Fig.~\ref{fig:ab_inversion} and Table~\ref{tab:ab_inversion}: (i) perturb all frames with random noise (Eqn.~\ref{eq:forward}), (ii) perturb all frames with same noise (Eqn.~\ref{eq:forward}), and (iii) DDIM inversion~\cite{song2021denoising}.
Compared to adding random noise, using DDIM inversion achieves better frame consistency and more continuous $x$-$t$ slice.
In row $1$ of Table~\ref{tab:ab_inversion}, albeit perturbing with same noise obtains higher frame consistency, it results in all frames becoming identical.

\noindent \textbf{Effect of sampling steps.}
VideoElevator uses more sampling steps than conventional T2V sampling (\ie, 100 steps vs. 50 steps), so concerns arise whether performance gains are due to increased sampling steps.
For a fair comparison, we evaluate the performance of typical T2V with $100$ steps in Table~\ref{tab:ab_steps}.
In row $1$ and $2$, the performance of T2V using $100$ and $50$ steps is similar, where the former shows slight improvements in frame consistency and quality only.
Meanwhile, VideoElevator still significantly outperforms conventional T2V using $100$ steps in terms of all metrics.
\section{Conclusion}
We introduce VideoElevator, a \textit{training-free} and \textit{play-and-plug} approach that boosts the performance of T2V baselines with versatile T2I.
VideoElevator explicitly decomposes each sampling step into two components, including temporal motion refining and spatial quality elevating.
Given noise latent at timestep $t$, temporal motion refining leverages a low-pass frequency filter to reduce its flickers, while applying T2V-based SDEdit to portray realistic motion.
When computing noise latent required by T2I, it performs deterministic inversion to preserve motion integrity.
Spatial quality elevating inflates the self-attention of T2I along the temporal axis, and directly harnesses inflated T2I to predict less noisy latent.
Extensive experiments demonstrate the effectiveness of VideoElevator under the combination of various T2V and T2I.
Even with foundational T2I, VideoElevator significantly improves T2V baselines in terms of text alignment, frame quality, and aesthetic score.
When integrating personalized T2I, it enables T2V baselines to faithfully produce various styles of videos.
%


%
%
\bibliographystyle{splncs04}
\bibliography{main}

\end{document}